\documentclass{article}
\usepackage{ijcai26}

\usepackage{times}
\usepackage{soul}
\usepackage{url}
\usepackage[hidelinks]{hyperref}
\usepackage[utf8]{inputenc}
\usepackage[small]{caption}
\usepackage{graphicx}
\usepackage{amsmath}
\usepackage{amsthm}
\usepackage{booktabs}
\usepackage{algorithm}
\usepackage{algorithmic}
\usepackage[switch]{lineno}

\usepackage{tikz}
\usepackage{color}
\title{Reasoning or Pattern Matching?
Probing Large Vision–Language Models with Visual Puzzles}

\author{
Maria Lymperaiou$^1$\and
Vasileios Karampinis$^1$\and
Giorgos Filandrianos$^1$\and
Angelos Vlachos$^1$\and
Chrysoula Zerva$^2$\And
Athanasios Voulodimos$^1$\\
\affiliations
$^1$National Technical University of Athens\\
$^2$Instituto de Telecomunicações, Lisbon, Portugal \\
\emails
\{marialymp, vkarampinis, geofila, aavlachos\}@ails.ece.ntua.com, chrysoula.zerva@tecnico.ulisboa.pt,
thanosv@mail.ntua.gr
}
\begin{document}

\maketitle

\begin{abstract}
Puzzles have long served as compact and revealing probes of human cognition, isolating abstraction, rule discovery, and systematic reasoning with minimal reliance on prior knowledge. Leveraging these properties, visual puzzles have recently emerged as a powerful diagnostic tool for evaluating the reasoning abilities of Large Vision–Language Models (LVLMs), offering controlled, verifiable alternatives to open-ended multimodal benchmarks. This survey provides a unified perspective of visual puzzle reasoning in LVLMs. We frame visual puzzles through a common abstraction and organize existing benchmarks by the reasoning mechanisms they target (inductive, analogical, algorithmic, deductive, and geometric/spatial), thereby linking puzzle design to the cognitive operations required for solving. Synthesizing empirical evidence across these categories, we identify consistent limitations in current models, including brittle generalization, tight entanglement between perception and reasoning, and a persistent gap between fluent explanations and faithful execution. By framing visual puzzles as diagnostic instruments rather than task formats, this survey elaborates on the state of LVLM reasoning and outlines key directions for future benchmarks and reasoning-aware multimodal systems.
\end{abstract}

\section{Introduction}
The surge of Large Vision-Language Models (LVLMs) has rapidly transformed the field of multimodal learning, establishing their widespread use on a variety of applications. This trajectory brings up a need for change in evaluation, requesting model diagnostics beyond perception and high-level recognition. A well-timed research concern questions whether LVLMs can actually \textit{reason} over perceived concepts, deeply comprehend hidden patterns and effortlessly generalize to unseen formulations. To this end, it remains uncertain whether such systems are capable of structure-aware reasoning or merely excel in complex pattern-matching.

%Large Vision-Language Models (LVLMs) have rapidly advanced the field of multimodal learning evolving into indispensable highly-intelligent  assistants. This trajectory has triggered a radical shift in evaluation: beyond measuring \emph{recognition}, current research directions focus on diagnosing \emph{reasoning}. In essence, a central open question is whether LVLMs can reliably comprehend latent patterns, follow internal states, and ultimately generalize beyond superficial correlations. So, it remains unclear to what extent current systems possess \emph{structure-sensitive} reasoning as opposed to sophisticated pattern matching supported by massive pretraining.

A seminal pitfall is that popular multimodal benchmarks often entangle reasoning with confounding factors, such as linguistic fluency, world knowledge, and dataset biases, allowing strong performance via ``shortcut strategies" rather than genuine visual reasoning \cite{cao2025visualcognitiongaphumans}. In these settings, failure is difficult to localize: `did the model fail to \emph{perceive} the input or fail to \emph{abstract} the rule?  This limitation complicates principled analysis of reasoning behavior in modern LVLMs, particularly under distribution shifts and compositional novelty.

To this end, we argue that \emph{visual puzzles} offer the most rigorous diagnostic instrument and a timely solution to the aforementioned evaluation challenge. Unlike open-ended tasks, puzzles form \emph{controlled} and \textit{verifiable} problem instances, where successful solutions hinge on extracting and manipulating visual structure under explicit or implicit constraints. Due to the isolated environments and controlled conditions visual puzzles impose, they allow us to decouple domain knowledge from pure logic, providing a uniquely sharp lens into the ``perception-reasoning gap". Thus, LVLM failure modes are easier to detect, interpret and compare.

%The recent surge of puzzle-based benchmarks signals a methodological shift toward \textbf{mechanism-oriented evaluation}. However, current benchmark proliferation risks fragmentation of the field, underscoring a \textit{unifying taxonomy} to connect puzzle designs to the specific reasoning capabilities they aim to measure, as well as to the failure modes they expose.
The rapid spread of visual puzzle benchmarks allows the transition to \textit{reasoning-oriented evaluation}, offering an appropriate closed-world structure to target specific model defects. However, those benchmarks appear in isolation, illustrating heterogeneous design specifications and evaluation criteria which limit their contribution in diagnosing the reasoning capabilities and failure modes of underlying models.
In this survey, we provide a systematic treatment of \textit{visual puzzle reasoning} for LVLMs, contributing to the following: 
\begin{itemize} \item We provide a unified abstraction for visual puzzles, integrating inputs, rule structures, and solution spaces. 
\item  We organize the literature into five core reasoning dimensions: \emph{inductive, analogical, algorithmic, deductive, and geometrical reasoning.} This reasoning-driven taxonomy enables a systematic comparison of model failures across superficially different tasks. 
\item By consolidating this rapidly evolving area, we synthesize empirical findings to expose recurring LVLM vulnerabilities and propose a roadmap for the next generation of more reasoning-aware LVLMs.
\end{itemize}

There exist a few recent surveys examining reasoning in LLMs and LVLMs, but they differ in scope and focus. \cite{giadikiaroglou-etal-2024-puzzle} consider textual puzzles only and do not address the visual modality at all. \cite{zhou2025perceptioncognitionsurveyvisionlanguage} study vision–language reasoning more broadly, analyzing general LVLM reasoning abilities without a specific focus on puzzle-based evaluation. Additionally, \cite{ke2025explainanswersurveycompositional} explore particular aspects of visual reasoning, such as compositionality, while \cite{Markaki2023JigsawSurvey} focus on narrow puzzle domains, such as jigsaw puzzle-solving. In contrast, we recast visual puzzles as diagnostic instruments for probing the reasoning abilities of LVLMs, offering a unified perspective that connects puzzle design, evaluation outcomes, and observed failure modes of LVLMs.

\section{Puzzle Definition and Taxonomy}
The survey focuses on a specific subset of reasoning tasks, which we refer to as \textit{visual puzzles}. A visual puzzle in the interest of this work holds the following core properties:
\begin{enumerate}
    \item Essential reliance on visual information.
    \item Presence of an explicitly constrained and verifiable puzzle structure that features latent rules, transformations, or relational dependencies within a constrained problem space, targeting the evaluation of cognitive abilities.
    \item Reasoning-centric challenges from the LVLM perspective, without reliance on knowledge priors.
\end{enumerate}

Based on these, we formalize a visual puzzle to be a problem instance  $\langle I,R,S\rangle$, where: the input $I$ refers to an information artifact concerning the visual input $\mathcal{V}$ and optional textual context $\mathcal{T}$; the rules $R$ concern \textit{explicit} (stated) or \textit{implicit} (to be inferred from $I$) constraints that govern valid solutions; the solution space
$S$ constitutes a typically discrete or combinatorial structured set of candidates, which must also be \textit{verifiable} via a ground-truth logical consistency check.

Our taxonomy considers the following categories:
\begin{enumerate}
    \item \textbf{Inductive Reasoning}: the process of inferring latent functions from a few inputs $I$ and generalizing them to novel instances. The governing principles are not explicitly provided, posing the major challenge of distinguishing abstractive capabilities of LVLMs from superficial pattern-matching. The rule set \textit{R} is \textit{implicit}, and if properly inferred (and consequently executed) leads to a \textit{deterministic} solution space $S$. 
    \item \textbf{Analogical Reasoning}: identifying relational correspondences between entities or situations and transferring those relationships to new contexts. The emphasis is on relational similarity and contrasting features that drive conceptual formulations rather than procedural execution or rule induction. Discovering analogies leads to an \textit{implicit} rule set \textit{R}. However, if multiple plausible analogies exist, applying these rules can yield diverging results, creating a \textit{non-deterministic} solution space \textit{S}.
    \item \textbf{Algorithmic Reasoning}: executing a sequence of operations or state transitions to reach a goal. The challenge lies in maintaining and updating an internal state across multiple steps rather than discovering latent rules, requiring multi-step reasoning capabilities of the model at hand. The rule set $R$ is \textit{explicit} and typically provided as problem instructions, the correct execution of which culminates in a \textit{deterministic} solution space $S$.
    \item \textbf{Deductive Reasoning}: deriving logically necessary conclusions from clearly stated premises or constraints, requiring constraint propagation and logical entailment capabilities from the solver. The corresponding rule set $R$ is \textit{explicit}, leading to \textit{deterministic} solutions $S$.
    \item \textbf{Geometric / Spatial Reasoning}: concerns spatial relationships, transformations, and structural configurations, challenging the model's ``mental imagery". It often requires conceptual rotation, spatial abstraction, or geometric invariance reasoning, which may either correspond to \textit{implicit or explicit} rule sets $R$ based on whether the task input is explicitly given or must be inferred; correspondingly, the degree to which $S$ is deterministic depends on the extent to which $R$ constrains the instance.
\end{enumerate}
The proposed taxonomy establishes a direct link to distinct cognitive operations leveraged in problem-solving, with the categories considered recurring frequently in respective benchmark literature. The dominant reasoning abilities exercised in visual puzzles confirm the diagnostic clarity that this taxonomy offers; in this sense, condensing categories would obscure distinct cognitive cues, whereas considering more fine-grained divisions would unreasonably fragment closely related findings. Hence, these puzzle categories are not mutually exclusive; they isolate dominant reasoning mechanisms that benchmarks emphasize, enabling proper comparison across seemingly different puzzle formats.

% taxonomy figure
%\color{red} {\huge figure:} \textbf{check here \url{https://www.canva.com/design/DAG96EmX0YA/_wW-hppdCvLNXu3PJ1UU0Q/view?utm_content=DAG96EmX0YA&utm_campaign=designshare&utm_medium=link2&utm_source=uniquelinks&utlId=h7c7453936c} to edit check here: \url{https://www.canva.com/design/DAG96EmX0YA/jYKXUNIqDPjrCKNXEaTU4w/edit?utm_content=DAG96EmX0YA&utm_campaign=designshare&utm_medium=link2&utm_source=sharebutton}} \color{black}
% taxonomy figure

\section{Benchmarks and Analysis}
\subsection{Inductive Reasoning}
Research on visual reasoning long predates LVLMs and has historically aimed to isolate the cognitive operations required to infer structure from visual inputs. A foundational paradigm for evaluating \emph{abstract visual pattern induction} is Raven’s Progressive Matrices (RPM) \cite{rpm}, where solvers complete a visual matrix utilizing patterns that involve shape, size and colors. Proceeding from 
measuring human cognition to evaluating neural network generalizability,  Procedurally Generated Matrices (PGM) \cite{pgm} were introduced as a precisely controlled framework of adjustable difficulty and systematic rule composition, following the RPM paradigm.
While PGM- and RPM-style tasks  have been crucial for studying abstraction in neural models, they typically operate over a relatively \emph{closed} attribute vocabulary (e.g., color, shape, position) with carefully factorized generative processes, and thus offer limited challenge for LVLMs.

\subsubsection{LVLM Benchmarks}
\paragraph{Foundational benchmarks}
With the emergence of LVLMs, inductive reasoning benchmarks have been revisited to probe their rule-discovery capabilities. The Abstraction and Reasoning Corpus (ARC) \cite{arc} introduces grid-based puzzles where few input-output exemplars represent an implied rule that should be generalized over new inputs. Contrary to RPM-based puzzles, ARC requires open-world assumptions, placing higher demands on generalization and abstraction. Consequently, EasyARC \cite{easyarc} simplifies visual input to factor out perceptual shortcomings so that reasoning is isolated. A competitive successor, ARC-AGI-2 \cite{arc-agi-2}, elevates the deliberate thinking needed even from the human perspective, probing complex reasoning cues and compositionality. In contrast to these high-level abstraction probes, BabyVision \cite{chen2026babyvisionvisualreasoninglanguage} targets the developmental roots of induction, establishing a baseline for early vision scenarios.

\paragraph{Knowledge-Decoupled Probes}
Several recent benchmarks are designed to disregard the usage of world knowledge and linguistic cues, enabling a more direct assessment of pure induction. VisualPuzzles \cite{visualpuzzles} serves as one of the most representative designs of abstracting away knowledge priors, while VisuRiddles \cite{visuriddles} addresses fine-grained perception as a bottleneck for appropriate induction. In a complementary direction, MME-Reasoning \cite{mme-reasoning} 
focuses on reasoning rather than perceptual abilities, again enforcing disentanglement from domain knowledge. 
PuzzleVQA \cite{puzzlevqa} formulates visual puzzle reasoning in a VQA format focusing on simplicity, diversity and reasoning auditing based on generated rationales.
VisualSphinx \cite{visualsphinx} targets large-scale LVLM training, with the goal of fusing inductive abilities which may not be adequately present by default. In a slightly different direction, LogicVista \cite{logicvista}, occasionally receiving hints regarding rule induction, prioritizes the evaluation of rule grounding and execution over mere rule discovery.

\paragraph{Generalization Probes}
Raven-inspired benchmarks further test whether induced abstractions transfer beyond the training distribution. Building on rule-annotated RPM extensions, such as I-RAVEN \cite{i-raven},  consequent works including A-I-RAVEN and I-RAVEN-Mesh \cite{ai-raven-mesh} stress generalization and transferability of inductive abilities to unseen configurations, attributes and visual renderings, while also challenging compositionality.

\paragraph{Human-Based Intelligence}
A line of work collates LVLM reasoning abilities to human intelligence using IQ-style artifacts. IQBench \cite{iqbench} directly employs IQ tests to assess LVLM reasoning paths and intermediate explanations, while MANBench \cite{manbench} probes induction in mazes and shape-based pattern matching.

\subsubsection{Analysis and Evaluation}
Synthesizing findings across benchmarks and diagnostic work, LVLMs exhibit systematic weaknesses that extend beyond accuracy and become most salient under alternative setups. A consistent finding is that failures arise from both fine-grained visual perception and the subsequent mapping to abstract variables, suggesting dividing analysis accordingly. 

\paragraph{Perceptual Limitations} Attribution of performance degradation is confused due to perceptual limitations \cite{easyarc} and modality-related particularities \cite{Opielka2024DoLL,zhang2025thinkvisuallyreasontextually} that obscure real inductive  abilities of LVLMs. When dissecting knowledge from perception, substantial performance drops  reveal the LVLMs' tendency to consult knowledge priors over visual inputs \cite{arc,visualpuzzles,mme-reasoning}. 
Furthermore, perturbations on non-essential visual factors can trigger sharp accuracy degradations  \cite{cao2025visualcognitiongaphumans}, suggesting reliance on superficial cues rather than invariant rule induction.  Such findings are verified in settings where LVLMs are tasked to transfer inferred rules to novel renderings \cite{ravenx,ai-raven-mesh}, outlining brittle out-of-distribution generalization. This limitation is further analyzed by \cite{chen2026babyvisionvisualreasoninglanguage}, revealing that LVLMs frequently conflate style with structure, leading to failures even in atomic inductive tasks solvable by young children. Comparison against human baselines demonstrates that LVLMs struggle with rule inference, instead favoring familiar patterns \cite{iqbench,manbench}.

\paragraph{The Perception-Reasoning Bottleneck}
Even when LVLMs succeed in perception, they fail to synthesize the underlying rule or generalize it to new instances. Model scaling and prompting are not sufficient to elicit induction \cite{visualpuzzles}, shifting the weight towards training-heavy solutions \cite{visualsphinx}. The need for post-training suggests a possibly inherent LVLM inability for inductive thinking; this shortcoming is further supported by the evidenced \textit{explanation-execution gap}: even though Chain of Thought (CoT) improves verbalization of thinking, the generated claims regarding inferred rules may be invalid given the visual evidence. \cite{chen2026babyvisionvisualreasoninglanguage} identify this as a ``verbalization bottleneck", where models attempt to approximate visual states through language, inevitably losing the fidelity required for robust induction. The discrepancy between fluency and reasoning fidelity indicates that plausible intermediate generations do not guarantee inductive comprehension and may even be misleading
\cite{visuriddles}. Generally, as long as the LVLM properly perceives inputs and infers the underlying rule, synthesizing the final solution is typically successful \cite{puzzlevqa}.

\subsection{Analogical Reasoning}
A foundational formulation of analogical reasoning is Bongard Problems \cite{Bongard1970PatternRecognition}: given two sets of images (positive and negative examples), the solver must infer the abstract concept that distinguishes them. Solving such problems does not involve executing an explicit procedure, but rather performing a comparative inference over visual relations to identify a shared conceptual distinction in a contrastive way.

\subsubsection{LVLM Benchmarks}
\paragraph{Contrastive Concept Discovery}
Bongard-style problems have been revisited in the LVLM era as a diagnostic tool for relational and conceptual abstraction, evolving from synthetic primitives to increasingly realistic visual contexts. Early adaptations focus on scaling semantic and visual richness: Bongard-HOI \cite{boingard-hoi} introduces human-object interactions to probe relational perception in natural images, while Bongard-OpenWorld \cite{bongardopenworld} relaxes closed-world assumptions by introducing open-vocabulary visual concepts. More recent benchmarks emphasize diagnostic precision. Bongard in Wonderland \cite{bongard-in-wonderland} revisits the original paradigm to diagnose failures in elementary abstract concepts despite their visual simplicity. Bongard-RWR+ \cite{bongardpwr} complements this direction by grounding fine-grained abstract concepts in real-world imagery, enabling controlled comparison between synthetic formulations and their natural visual counterparts.

\paragraph{Relational and Structural Mapping}
A complementary line of benchmarks targets analogical reasoning as the ability to map relational structure across visual representations. Moving beyond literal symbol interpretations,  REBUS \cite{rebus} and COLUMBUS \cite{columbus} benchmarks evaluate LVLMs' ability to capture visual meanings and semantically map them to language through wordplay.  MARVEL \cite{marvel} detaches reasoning from language understanding, introducing geometric or abstract shapes as context for testing relational alignment. Moreover, it decreases knowledge demands and disconnects visual perception to assess structural mapping abilities in isolation. In a similar sense, VisualPuzzles \cite{visualpuzzles} evaluates analogical mapping in knowledge-light settings of varying difficulty, exposing inherent reasoning failures. VOILA \cite{voila} extends this line of work by increasing the perceptual load and explicitly coupling analogical reasoning with perceptual understanding.

\subsubsection{Analysis and Evaluation}
Diagnostic analyses across analogical benchmarks showcase that LVLMs tend to favor superficial visual attributes over relational structures, leading to brittle analogical behavior.

\paragraph{Perceptual Bias}
A dominant failure mode of LVLMs concerns an over-reliance on individual visual inputs rather than discovering relational isomorphisms. Spurious cues in Bongard-style puzzles (color, texture, or object cardinality) override LVLMs' perception of intended analogies, leading to sharp performance degradation, especially under minor variations or controlled distribution shifts \cite{małkiński2025reasoninglimitationsmultimodallarge,bongard-in-wonderland}. Analyses on VisualPuzzles rationales confirm LVLMs' adherence on high-level visual features rather than contextualizing the transformation needed to solve the analogy \cite{visualpuzzles}.
These findings denote a limited ability of performing comparative inference over superficial pattern-matching. Similar evidence from Rebus puzzles \cite{rebus-puzzles,columbus} concludes that LVLMs default to literal symbolic descriptions, failing to perform the required relational mapping to language.

\paragraph{The Relational Mapping Deficit}
Even when the initial perceptual step is executed as intended, consequent reasoning steps fail to preserve relational alignment. MARVEL \cite{marvel} exposes LVLMs' failure in performing the analogy step despite their capacity on perceiving visual elements in context, as occurring by their suggested evaluation breakdown between representation and mapping. VOILA \cite{voila} reinforces this observation, reporting weak transfer across instances on explicit analogical reasoning tasks even for strong LVLMs, indicating limited generalization.

%Collectively, these results point to a persistent grounding-mapping bottleneck: abstract relations inferred from a source configuration are not stably preserved when applied to a target configuration. This indicates that current LVLM architectures struggle to hold and manipulate relational representations across distinct visual contexts.

%Recent benchmarks target algorithmic reasoning through visually grounded puzzles and games that require procedural execution. PuzzleBench provides a dynamic evaluation framework spanning diverse puzzle types with multi-step solution paths, enabling controlled assessment of rule execution and state tracking in multimodal settings \cite{puzzlebench}. PUZZLEPLEX emphasizes reasoning and planning over puzzle environments, requiring models to solve structured puzzles through sequences of decisions under explicit constraints \cite{puzzleplex}.
\subsection{Algorithmic Reasoning}
Contrary to the previous categories, algorithmic reasoning lacks a foundational pre-LVLM paradigm, with origins of the required puzzle-solving abilities encountered in Jigsaw puzzles, LEGO assembly and other procedural rule-based games.

\subsubsection{LVLM Benchmarks}
\paragraph{Static symbolic puzzles}
A class of benchmarks targets discrete algorithmic reasoning grounded in visual grids or structured layouts, enabling evaluation of LVLMs' ability on different scales of procedural depth and constraint complexity. PUZZLES \cite{puzzles} comprises a variety of programmatically generated puzzles with diverging problem size and difficulty, suited to test generalization abilities of LVLMs in algorithmic execution over a well-defined action space. AlgoPuzzleVQA \cite{algopuzzlevqa} also considers grid-based puzzles that target action optimization, combinatorics, search logic, as well as multimodal arithmetic and graph reasoning framed in a VQA format. In a similar sense, VisualPuzzles algorithmic instances \cite{visualpuzzles} emphasize procedural execution, including perceptual constraint satisfaction, combinatorics and multimodal arithmetic.

\paragraph{Spatially Procedural Puzzles}
A couple of algorithmic benchmarks such as Jigsaw-Puzzles \cite{jigsaw-puzzles} and LEGO-Puzzles \cite{lego-puzzles} move beyond static one-shot recognition, assessing procedural reasoning through multi-step spatial manipulation under physical or geometric constraints. Even though the problem specification is available upfront, solutions require maintaining and updating accumulating spatial representations during procedural construction steps, challenging LVLMs' ability of sustaining coherent internal transformations in each execution step.

\paragraph{Interactive Planning}
A third category of benchmarks aims to evaluate whether LVLMs can execute explicit procedures over visual environments whose state dynamically evolves in response to sequential actions, questioning LVLMs' ability to preserve consistent internal states across perception-action cycles. PuzzleBench \cite{puzzlebench} introduces an interactive puzzle environment, in which LVLMs are tasked to decide on sequential actions that lead to updating verifiable intermediate steps. ING-VP \cite{ingvp} promotes simple, minimally interactive games to disentangle perceptual challenges, such as comprehending visual concepts or rules, from actual action implementation and planning, facilitating attribution of LVLM failures. BALROG \cite{balrog} regards reasoning as acting within a dynamic environment with delayed rewards, posing a special focus on agentic interaction over long-horizon decision-making.
To further stress model planning abilities, PuzzlePlex \cite{puzzleplex} focuses on multi-stage interactive execution in both deterministic and stochastic settings, targeting LVLM reasoning and strategic coherence under evolving constraints. Further broadening the evaluation scope, ENIGMAEVAL \cite{enigmaeval} introduces highly difficult multimodal challenges of varying input format with unambiguous solutions, pinpointing the limits of LVLMs' procedural cognition.

\subsubsection{Analysis and Evaluation}
Across algorithmic benchmarks, LVLMs exhibit systematic weaknesses in executing explicit procedures over extended horizons, even when task rules are clearly specified.

\paragraph{The Perception-Execution Gap}
Perceptual, understanding and reasoning errors serve as culprits for limited LVLM performance \cite{visualpuzzles}. Even though perceptual limitations account for a non-negligible bottleneck early on, actual execution during planning and state maintenance arise as central limitations when multi-step reasoning and sequential actions are required \cite{ingvp,puzzlebench,lego-puzzles,jigsaw-puzzles}. This finding is further evidenced in interactive environments where decisions become progressively misaligned from visual states, resulting in logical derailment of LVLMs \cite{puzzleplex,balrog,enigmaeval}. Another indicator of the perception-execution gap is the disconnect between reasoning verbalization and actual execution: even though plausible step-by-step explanations are generated, the realized actions end up violating task rules or fail to incorporate the actual visual state \cite{vlm-game-players,ingvp}. This explanation-execution gap suggests that CoT-style prompting does not guarantee faithful state tracking or procedural consistency \cite{algopuzzlevqa}. 

\paragraph{Performance Degradation with Procedural Depth}
A recurring observation in both static and dynamic puzzles is that performance of LVLMs degrades sharply as solution length and state-tracking demands increase \cite{algopuzzlevqa,puzzles,puzzlebench}, showcasing that maintaining long-horizon sequential coherency poses a significant challenge. This finding is confirmed in spatial assembly tasks, where LVLMs often succeed on isolated or short-horizon steps but accumulate errors over longer sequences, leading to violated constraints or inconsistent intermediate configurations \cite{lego-puzzles,jigsaw-puzzles}.

\subsection{Deductive Reasoning}
As in the case of algorithmic reasoning, there are no canonical pre-LVLM benchmarks for deductive reasoning. Constraint-satisfaction visual puzzles, such as Sudoku, serve as the initial conceptual inspiration for later benchmarks.

\subsubsection{LVLM Benchmarks}

\paragraph{Symbolic Constraints}
A primary class of deductive benchmarks evaluates whether LVLMs can apply explicitly stated symbolic rules to visually grounded entities. LogicVista \cite{logicvista} enables deduction over facts extracted from images (diagram structure, spatial relationships etc) following formally specified rules, with the primary goal of binding visual elements to symbols that drive logical execution. VisuLogic \cite{visulogic} emphasizes visually grounded deduction by introducing hard-to-articulate visual instances to evaluate the role of visual comprehension without verbal conflating. Moreover, a set of benchmarks evaluates the application of deduction over intermediate structured representations extracted from visual inputs. VisualSphinx \cite{visualsphinx} frames deduction as a trait that can be potentially acquired during training, with post-training gains reflecting an LVLM's previously limited deductive ability. MME-Reasoning \cite{mme-reasoning} enforces minimal reliance on knowledge while reducing the perceptual burden to ensure separation of reasoning concerns, similar to the main goals of VisualPuzzles deductive puzzles \cite{visualpuzzles}. 

\paragraph{Grid-Based Constraints}
Deductive reasoning is also operationalized through grid-based visual puzzles, posing the additional cognitive load of obeying structured layouts during constraint propagation in order to deterministically reach a valid solution.
Sudoku-Bench \cite{sudokubench} evaluates whether LVLMs can solve Sudoku-style variants by maintaining global consistency while making local assignments under explicit rules. Unlike the fixed rules of Sudoku, VGRP-Bench \cite{vgrp-bench} introduces a diverse set of rules over grids of varying grid size and decision space, while also disentangling reasoning from visual perception. Finally, PUZZLES \cite{puzzles} positions grid-based deduction within a formal setup that allows testing constraint propagation during multi-step symbolic manipulation.

%Related grid-based logic tasks, including Latin Square-based puzzles (like Sudoku) and Binairo variants in VGRP-Bench \cite{vgrp-bench}, similarly test whether models can deduce the state of a cell based on fixed neighborhood constraints. PUZZLEVQA \cite{puzzlevqa} enables controlled evaluation by embedding abstract pattern reasoning in a question-answering framework.  The intended difficulty across these settings lies in synthesizing a coherent transformation rule $R$ from locally interpretable primitives. These benchmarks emphasize deterministic reasoning, where valid solutions are uniquely determined by correct constraint integration.

\subsubsection{Analysis and Evaluation}
\paragraph{Error Propagation from Visual Grounding}
Deductive failures are further amplified by the interaction between perception and logic. Perceptual failures often become evident in consequent stages, since error cascading from slight misperceptions progressively disturbs constraint satisfaction \cite{sudokubench}. Visual grounding presents an important bottleneck for downstream performance, highlighting the difficulty of maintaining consistent bindings between visual input and logical variables \cite{zhang2024how}. This is further confirmed when verbalization of the visual problem is excluded \cite{visulogic}, therefore purely blaming the visual modality for cascading failures.
Error decomposition confirms misinterpretations of grid structure and complex visual components, and reveals failures to prioritize actions or backtrack, contributing further to reasoning decay \cite{vgrp-bench}.

% justifying reasoning decay more thoroughly \cite{vgrp-bench}.

\paragraph{Constraint Integration Limitations}
Across deductive benchmarks, a recurring limitation is the difficulty in jointly propagating multiple constraints, even when all rules are explicitly provided. While LVLMs may apply individual constraints correctly, they often fail to integrate them into a globally consistent solution, resulting in partially valid or internally inconsistent outputs. This behavior is consistently observed across formal logic tasks and grid-based puzzles \cite{logicvista,visulogic,zhang2024how}. Moreover, a persistent gap emerges between verbalized explanations and deductive correctness: models frequently generate coherent-looking rationales that do not correspond to valid entailment or consistent constraint propagation, indicating that linguistic fluency is not a reliable indicator of deductive accuracy.

\subsection{Geometric/Spatial Reasoning}
Geometric and spatial reasoning is an \textit{orthogonal category} to the rest and rarely appears in isolation. Instead, it further stress tests the models to reason over spatial and geometric configurations together with implicit rule extraction, procedural execution or constraint satisfaction. Thus, it aims to discover model weaknesses attributed to grounding on respective concepts, represent them in consequent reasoning stages and combine them with aforementioned reasoning demands. Classical geometrical problems, mental rotations and construction puzzles form some pre-LVLM data inspirations.

%increasing task difficulty by requiring models to reason over spatial structure, invariance, and transformation in conjunction with abstraction, procedural execution, or logical constraints. As a result, geometric reasoning often acts as a stress factor that amplifies weaknesses in visual grounding, representation stability, and transformation consistency.

\subsubsection{LVLM Benchmarks}

\paragraph{Geometric Transformations in Inductive Puzzles}
A class of benchmarks embeds geometric reasoning directly into inductive rule discovery. Grid-based and matrix-style puzzles often require inferring transformations (rotation, reflection, symmetry, scaling, spatial repetition) as part of the latent rule. Raven-inspired extensions such as A-I-RAVEN and I-RAVEN-Mesh \cite{ai-raven-mesh} explicitly incorporate geometric attributes into abstract pattern induction, enabling controlled evaluation of whether models can generalize geometric relations beyond surface appearance. Complementarily, BabyVision \cite{chen2026babyvisionvisualreasoninglanguage} exposes that this inductive fragility extends to geometric primitives, showing that LVLMs struggle to induce simple rotation or reflection rules when distracted by other basic attributes. Large-scale abstract reasoning benchmarks such as VisualPuzzles and VisuRiddles \cite{visualpuzzles,visuriddles} further emphasize that successful induction often hinges on fine-grained spatial perception and the ability to generalize geometric regularities across visually diverse instances.

\paragraph{Spatial Structure in Analogical Reasoning}
In analogical reasoning puzzles, spatial relations mostly serve as the carrier of conceptual distinction. Bongard-style puzzles often encode abstract concepts through geometric configurations (relative position, enclosure, alignment, symmetry) \cite{bongard-in-wonderland,bongardpwr}. In these settings, spatial structure provides the basis for contrastive comparison, requiring models to identify relational correspondences rather than matching individual shapes. Benchmarks such as VOILA \cite{voila} explicitly probe this interaction by evaluating whether models can preserve spatial relations when transferring analogical structure across instances, isolating spatial alignment as a core component of analogy formation.

\paragraph{Spatial State Tracking in Procedural Tasks}
Algorithmic puzzles often impose explicit rules over spatially structured environments, involving assembly, navigation, or game-like interaction. Jigsaw-Puzzles and LEGO-Puzzles \cite{jigsaw-puzzles,lego-puzzles} are designed to evaluate whether LVLMs can track part configurations and predict spatial transformations across intermediate steps. Similarly, BALROG \cite{balrog} embeds spatial reasoning into agentic interaction, where correct action selection depends on preserving a coherent spatial representation over long horizons. These benchmarks focus on spatial state consistency as a prerequisite for reliable multi-step procedural execution.  

\paragraph{Spatial Grounding in Deductive Constraints}
Deductive reasoning benchmarks often rely on geometric structure to define the domain over which logical constraints are applied. Visual logic tasks require models to interpret spatial layouts, diagrams, or grids as structured representations of entities and relations. Across grid-based benchmarks such as VGRP-Bench, PUZZLES \cite{vgrp-bench,puzzles} and symbol-based benchmarks \cite{visulogic,logicvista}, spatial configuration determines which constraints apply and how they propagate. Here, geometric reasoning is not auxiliary but foundational: correct deduction depends on accurate spatial grounding of symbolic variables.

\paragraph{Geometry-Centric Reasoning Benchmarks}
A smaller set of benchmarks places geometric reasoning at the core of the task formulation. GeoSketch \cite{geosketch} focuses on explicit geometric construction and transformation, requiring models to reason over rotations, reflections, affine mappings, and auxiliary line constructions. These tasks probe whether LVLMs can maintain geometric consistency across intermediate constructions rather than relying on visual similarity. Complementary work on structured spatial problem solving, such as Rubik’s Cube–style tasks \cite{structured-rubik}, evaluates whether models can decompose complex geometric configurations into modular subproblems and coordinate sequential spatial operations under strict constraints.

\subsubsection{Analysis and Evaluation}
Across benchmark families, geometric and spatial reasoning consistently amplifies LVLM failure modes, revealing limitations that are less apparent in non-spatial settings.

\paragraph{Sensitivity to Superficial Spatial Cues}
In inductive and analogical benchmarks, LVLMs often rely on superficial spatial features, such as absolute position, local alignment, or visual similarity, rather than abstract geometric relations.
As a result, generalization degrades under transformations or compositional changes, even when the geometric rule is simple \cite{visuriddles,bongard-in-wonderland,chen2026babyvisionvisualreasoninglanguage}. Early spatial parsing errors often prevent abstraction.

%As a result, generalization degrades sharply under transformations or compositional changes, even when the underlying geometric rule is simple \cite{visuriddles,bongard-in-wonderland,chen2026babyvisionvisualreasoninglanguage}. . 
% Incorrect spatial parsing at early stages frequently prevents successful abstraction altogether.

\paragraph{Error Accumulation}
Spatial reasoning failures appear as error accumulation over consequent stages of execution. Minor inaccuracies in spatial interpretation propagate across steps, leading to invariant violations or inconsistent intermediate states \cite{lego-puzzles,jigsaw-puzzles}. These effects are pronounced in interactive or long-horizon algorithmic settings, where maintaining a stable spatial representation is essential for proper execution \cite{balrog}.

\paragraph{Geometry as a Bottleneck for Deduction}
Deductive benchmarks further show that spatial misinterpretation is a major source of logical failure. LVLMs often violate global consistency not because of incorrect rule application, but due to misaligned or misidentified spatial relations within the visual structure \cite{zhang2024how,visulogic}. This coupling between geometry and logic highlights the difficulty of separating spatial understanding from higher-level inference.

\paragraph{Fluent Explanations without Spatial Fidelity}
Finally, diagnostic evaluations indicate that geometric reasoning errors are predominantly masked by fluent but unfaithful explanations, positioning geometric and spatial reasoning as a stringent stress test for LVLMs. Although LVLMs may produce plausible verbal justifications, they frequently do not correspond to the actual spatial evidence used to reach a conclusion \cite{verify,lego-puzzles,visuriddles}. Such behaviors obscure underlying grounding failures and expose gaps between apparent reasoning competence and genuine, structure-preserving inference.

\section{Discussion and Future Directions}
Our analysis across different reasoning categories indicates a consistent LVLM performance drop in constrained and verifiable reasoning setups, despite their confirmed linguistic fluency and general knowledge, which may obscure deeper cognitive pitfalls. This core finding highlights potential causes and outlines significant future research directions.
%The progression from descriptive recognition to structure-sensitive reasoning marks a critical inflection point for LVLMs. By surveying visual puzzles across inductive, analogical, algorithmic, deductive, and geometric reasoning, this work exposes a persistent competence--performance gap: while LVLMs often exhibit linguistic fluency suggestive of reasoning, they frequently fail to execute rules faithfully when faced with controlled, checkable visual structures. In this section, we synthesize these findings and outline directions for advancing reasoning-oriented multimodal research.

\subsection{The Illusion of Reasoning}
A major insight occurring from this survey is the discrepancy between visual perception and concept manipulation during thinking. Error cascading stemming from small perceptual failures, such as misidentification of objects, erroneous counting or misperceived relationships derails downstream reasoning, even though a seemingly rational thinking procedure is demonstrated.
This bottleneck manifests as an \textit{illusion of reasoning}: LVLMs generate sound intermediate explanations, which however do not always correspond to explicit visual inputs; such discrepancies are especially noticeable in the highly restricted puzzle settings, verifying their choice as ideal diagnostic testbeds. Similar findings have been reported in LVLM reasoning on commonsense data \cite{zhang2025mmcotabenchmarkprobingvisual} and in language-only reasoning models, where fluent explanations obscure models' collapse in increasingly complex problems \cite{illusion}. Nevertheless, reasoning fidelity in visual puzzles is still in its infancy \cite{verify,voila}, suggesting a crucial future research avenue to be explored.
The illusion of reasoning suggests that intermediate explanations should be verified in context, particularly with grounding to visual inputs and instructions, inferred rules and intermediate states, while employed evaluation frameworks should be revisited to account for explanation faithfulness beyond mere performance-driven concerns.
%Such failures are especially visible in puzzles with deterministic solution spaces, where errors cannot be attributed to perceived ambiguity. These observations caution against equating explanation quality with reasoning fidelity and highlight the diagnostic value of visual puzzles, which expose grounding failures more reliably than open-ended multimodal tasks.

\subsection{Multi-Agent Collaboration}
Agentic structures arise as a promising direction, following their widespread adoption in language-only tasks. Specialized agents on distinct roles which collaborate and debate are able to eliminate reasoning errors and increase overall performance and robustness \cite{debate,liang-etal-2024-encouraging}. By offering multiple perspectives, LVLM agents can efficiently discuss over puzzle structure, identify candidate rules and verify that constraints are satisfied, rejecting erroneous hypotheses via interaction and diversification. From an interpretability perspective, agents can augment the reasoning traces provided through their consensus and disagreement verbalization, offer uncertainty insights and disentangle different types of failure modes of constituent LVLMs. 
To this end, agentic structures can help diminish the illusion of reasoning by employing an explanation judge that verifies the explanation accurately satisfies the input and constraints.

% Agentic structures can help diminish the illusion of reasoning, employing an explanation judge responsible for verifying that the explanation satisfies the input and the constraints.

\subsection{Revisiting Training Objectives}
Algorithmic and geometric puzzles reveal a pronounced complexity cliff for LVLMs: even though models perform adequately in static or short-horizon setups, they degrade sharply as the solution length increases. This attributes the root causes not on the lack of knowledge on the task, but in LVLMs' inability to maintain long reasoning traces and verify each intermediate step. The currently prevalent training objective of \textit{next token prediction} rewards statistical pattern matching and fluent language rather than reasoning perseverance in long-dependency setups, offering limited stimulus for state-consistent reasoning chains. Long-horizon training objectives have been proposed for language-only reasoning \cite{mahajan2025multitokenpredictionpretrainingllms}, suggesting a promising solution for improving LVLM reasoning as well. Moving one step further, future training objectives should consider intermediate state consistency and penalize violations, as well as take into account counterfactual invariance for advanced robustness.

%As solution length grows or spatial configurations evolve over multiple steps, performance degrades sharply. This pattern is not indicative of missing knowledge, but of insufficient computational persistence: internal representations of the latent state deteriorate as reasoning unfolds. These findings suggest limits of purely next-token prediction objectives for tasks requiring long-horizon state maintenance. Future research should explore process-oriented training signals, such as supervision over intermediate states, explicit reasoning traces, or verification-based objectives that reward consistency across steps. Iterative refinement loops, where models propose, check, and revise intermediate solutions, offer a promising direction for bridging the gap between symbolic reasoning and faithful execution.

\subsection{Towards Compositional Generalization}
Superficial visual characteristics, such as color, texture or style tend to be overly influential to LVLM decision-making, a prevalent observation across inductive and analogical benchmarks. Especially when LVLMs are tasked to operate in non-familiar puzzle formulations, such as those involving multiple reasoning stages and novel compositions, the memorized patterns cannot further serve as shortcuts, leading to inevitable performance collapse. Currently, compositionality is rarely treated as a design priority for visual puzzles and mostly appears as a byproduct of dedicated needs tailored to each reasoning type. This suggests that dataset design should consider puzzle setups with well-defined compositional probes, such as definite compositional depth and breadth, disentanglement of individual variables and controlled variations to enhance abstraction and generalization.

%Across inductive and analogical benchmarks, LVLMs exhibit strong sensitivity to non-essential visual factors such as texture, color, or rendering style. When familiar rules are applied to novel primitives or geometric transformations, performance often collapses. This suggests that many LVLMs rely on surface-level templates rather than invariant abstractions. 
%Achieving structure-sensitive reasoning will require both benchmarks and training regimes that explicitly stress compositional generalization. Synthetic puzzle environments where visual attributes and logical rules are systematically decoupled play a crucial role here. By minimizing semantic shortcuts and enforcing controlled variation, such benchmarks discourage pattern memorization and more directly probe abstraction.

\subsection{Abductive and Lateral Visual Reasoning}
A notable gap identified by this survey is the near absence of \textit{hypothesis-driven} puzzle-solving as instantiated by \textit{abductive} and \textit{lateral} reasoning on visual puzzles. These reasoning forms entail more advanced cognitive demands, requiring inference under uncertainty and incomplete information, examination of multiple plausible hypotheses and overriding presuppositions induced from default reasoning pathways. Such demands have been recently addressed in purely linguistic benchmarks \cite{lateval,Todd2024MissedCL,jiang-etal-2024-semeval,dougrez-lewis-etal-2025-assessing},  indicating that model capacity is far more limited in uncertain formulations. The sparse   literature in visual benchmarks \cite{mme-reasoning,rebus} suggests that extending lateral and abductive thinking accordingly serves as a natural and important future direction beyond rule-based execution.  Such endeavors will enable the assessment of hypothesis revision, reasoning over partial or ambiguous evidence and verification of multiple thinking branches for LVLMs.

\section{Conclusion}
This survey proposes the adoption of visual puzzles as an evaluation framework for LVLM reasoning. By explicitly defining visual puzzles and dividing them according to the primary required reasoning mechanism, we outline a connection with cognitive operations, including inductive, analogical, algorithmic, deductive, and geometric reasoning types. The constrained and verifiable nature of puzzles eliminates ambiguity in evaluation or reliance on knowledge priors, rendering puzzles as a pure logical reasoning evaluator.
Across reasoning categories, LVLMs display a notable gap between fluent explanations and accurate logical execution, diminished conceptual implementation in comparison to perceptual pattern understanding and brittle generalization to novel instantiations. As a result, LVLMs' apparent reasoning competence can be mostly attributed to successful pattern matching rather than faithful inference, raising concerns regarding LVLMs inherent abilities in abstraction, state tracking or adherence to constraints. Overall, puzzle-driven findings highlight the development of more reasoning-aware, trustworthy LVLMs that operate beyond shortcut-driven behavior.

%This survey positions visual puzzles as a principled framework for probing several reasoning dimensions in LVLMs. By organizing benchmarks around the reasoning mechanisms they require, we connected puzzle design to the cognitive operations and failure modes exhibited by current systems. Across inductive, analogical, algorithmic, deductive, and geometric settings, LVLMs display a persistent gap between fluent explanations and faithful execution, brittle generalization, and strong dependence on precise visual grounding. These findings suggest that apparent reasoning competence often reflects pattern matching rather than structure-sensitive inference. Beyond evaluation, visual puzzles offer a controlled testbed for studying abstraction, state tracking, and rule execution under verifiable conditions. As such, they provide both a diagnostic lens and a foundation for developing reasoning-aware LVLMs that move beyond shortcut-driven behavior.

\newpage  % this helps me see references budget
\bibliographystyle{named}
\bibliography{ijcai26}

\begin{thebibliography}{}

\bibitem[\protect\citeauthoryear{Barrett \bgroup \em et al.\egroup }{2018}]{pgm}
David Barrett, Felix Hill, et~al.
\newblock Measuring abstract reasoning in neural networks.
\newblock In {\em ICML}, volume~80, pages 511--520, 2018.

\bibitem[\protect\citeauthoryear{Bi \bgroup \em et al.\egroup }{2025}]{verify}
Jing Bi, Junjia Guo, et~al.
\newblock Verify: A benchmark of visual explanation and reasoning for investigating multimodal reasoning fidelity.
\newblock {\em arXiv:2503.11557}, 2025.

\bibitem[\protect\citeauthoryear{Bongard}{1970}]{Bongard1970PatternRecognition}
Mikhail~M. Bongard.
\newblock {\em Pattern Recognition}.
\newblock Spartan Books, New York, 1970.

\bibitem[\protect\citeauthoryear{Camposampiero \bgroup \em et al.\egroup }{2025}]{ravenx}
Giacomo Camposampiero, Michael Hersche, et~al.
\newblock {I-RAVEN-X: Benchmarking Generalization and Robustness of Analogical and Mathematical Reasoning in Large Language and Reasoning Models}.
\newblock 2025.

\bibitem[\protect\citeauthoryear{Cao \bgroup \em et al.\egroup }{2025}]{cao2025visualcognitiongaphumans}
Xu~Cao, Yifan Shen, et~al.
\newblock What is the visual cognition gap between humans and multimodal llms?
\newblock {\em arXiv:2406.10424}, 2025.

\bibitem[\protect\citeauthoryear{Chen \bgroup \em et al.\egroup }{2026}]{chen2026babyvisionvisualreasoninglanguage}
Liang Chen, Weichu Xie, et~al.
\newblock Babyvision: Visual reasoning beyond language.
\newblock {\em arXiv:2601.06521}, 2026.

\bibitem[\protect\citeauthoryear{Chia \bgroup \em et al.\egroup }{2024}]{puzzlevqa}
Yew~Ken Chia, Vernon Toh, et~al.
\newblock {P}uzzle{VQA}: Diagnosing multimodal reasoning challenges of language models with abstract visual patterns.
\newblock In {\em ACL Findings}, pages 16259--16273, 2024.

\bibitem[\protect\citeauthoryear{Chollet \bgroup \em et al.\egroup }{2025}]{arc-agi-2}
Francois Chollet, Mike Knoop, et~al.
\newblock Arc-agi-2: A new challenge for frontier ai reasoning systems.
\newblock {\em arXiv:2505.11831}, 2025.

\bibitem[\protect\citeauthoryear{Chollet}{2019}]{arc}
François Chollet.
\newblock On the measure of intelligence.
\newblock {\em arXiv:1911.01547}, 2019.

\bibitem[\protect\citeauthoryear{Das \bgroup \em et al.\egroup }{2025}]{rebus-puzzles}
Trishanu Das, Abhilash Nandy, et~al.
\newblock A large and diverse multimodal benchmark for evaluating the ability of vision-language models to understand rebus puzzles.
\newblock {\em arXiv:2511.01340}, 2025.

\bibitem[\protect\citeauthoryear{Dougrez-Lewis \bgroup \em et al.\egroup }{2025}]{dougrez-lewis-etal-2025-assessing}
John Dougrez-Lewis, Mahmud~Elahi Akhter, et~al.
\newblock Assessing the reasoning capabilities of {LLM}s in the context of evidence-based claim verification.
\newblock In {\em ACL Findings}, pages 20604--20628, July 2025.

\bibitem[\protect\citeauthoryear{Du \bgroup \em et al.\egroup }{2024}]{debate}
Yilun Du, Shuang Li, et~al.
\newblock Improving factuality and reasoning in language models through multiagent debate.
\newblock In {\em ICML}. JMLR.org, 2024.

\bibitem[\protect\citeauthoryear{Estermann \bgroup \em et al.\egroup }{2024}]{puzzles}
Benjamin Estermann, Luca~A. Lanzend\"{o}rfer, et~al.
\newblock Puzzles: a benchmark for neural algorithmic reasoning.
\newblock In {\em NeurIPS}, 2024.

\bibitem[\protect\citeauthoryear{Fan and Yuan}{2025}]{structured-rubik}
Chongshan Fan and Shenghai Yuan.
\newblock Structured task solving via modular embodied intelligence: A case study on rubik's cube.
\newblock {\em arXiv:2507.05607}, 2025.

\bibitem[\protect\citeauthoryear{Feng \bgroup \em et al.\egroup }{2025}]{visualsphinx}
Yichen Feng, Zhangchen Xu, et~al.
\newblock Visualsphinx: Large-scale synthetic vision logic puzzles for rl.
\newblock {\em arXiv:2505.23977}, 2025.

\bibitem[\protect\citeauthoryear{Ghosal \bgroup \em et al.\egroup }{2025}]{algopuzzlevqa}
Deepanway Ghosal, Vernon Toh, et~al.
\newblock {A}lgo{P}uzzle{VQA}: Diagnosing multimodal reasoning challenges of language models with algorithmic multimodal puzzles.
\newblock In {\em NAACL}, pages 9615--9632, 2025.

\bibitem[\protect\citeauthoryear{Giadikiaroglou \bgroup \em et al.\egroup }{2024}]{giadikiaroglou-etal-2024-puzzle}
Panagiotis Giadikiaroglou, Maria Lymperaiou, et~al.
\newblock Puzzle solving using reasoning of large language models: A survey.
\newblock In {\em EMNLP}, pages 11574--11591, 2024.

\bibitem[\protect\citeauthoryear{Gritsevskiy \bgroup \em et al.\egroup }{2024}]{rebus}
Andrew Gritsevskiy, Arjun Panickssery, et~al.
\newblock Rebus: A robust evaluation benchmark of understanding symbols.
\newblock {\em arXiv:2401.05604}, 2024.

\bibitem[\protect\citeauthoryear{Hu \bgroup \em et al.\egroup }{2021}]{i-raven}
Sheng Hu, Yuqing Ma, et~al.
\newblock Stratified rule-aware network for abstract visual reasoning.
\newblock In {\em AAAI}, volume~35, pages 1567--1574, 2021.

\bibitem[\protect\citeauthoryear{Huang \bgroup \em et al.\egroup }{2024}]{lateval}
Shulin Huang, Shirong Ma, et~al.
\newblock {L}at{E}val: An interactive {LLM}s evaluation benchmark with incomplete information from lateral thinking puzzles.
\newblock In {\em LREC-COLING 2024}, pages 10186--10197, 2024.

\bibitem[\protect\citeauthoryear{Jiang \bgroup \em et al.\egroup }{2022}]{boingard-hoi}
Huaizu Jiang, Xiaojian Ma, et~al.
\newblock Bongard-hoi: Benchmarking few-shot visual reasoning for human-object interactions.
\newblock In {\em CVPR}, 2022.

\bibitem[\protect\citeauthoryear{Jiang \bgroup \em et al.\egroup }{2024a}]{jiang-etal-2024-semeval}
Yifan Jiang, Filip Ilievski, and Kaixin Ma.
\newblock {S}em{E}val-2024 task 9: {BRAINTEASER}: A novel task defying common sense.
\newblock In {\em SemEval-2024}, pages 1994--2008, 2024.

\bibitem[\protect\citeauthoryear{Jiang \bgroup \em et al.\egroup }{2024b}]{marvel}
Yifan Jiang, Jiarui Zhang, et~al.
\newblock Marvel: Multidimensional abstraction and reasoning through visual evaluation and learning.
\newblock In {\em NeurIPS}, volume~37, pages 46567--46592, 2024.

\bibitem[\protect\citeauthoryear{Ke \bgroup \em et al.\egroup }{2025}]{ke2025explainanswersurveycompositional}
Fucai Ke, Joy Hsu, et~al.
\newblock Explain before you answer: A survey on compositional visual reasoning.
\newblock {\em arXiv:2508.17298}, 2025.

\bibitem[\protect\citeauthoryear{Kraaijveld \bgroup \em et al.\egroup }{2024}]{columbus}
Koen Kraaijveld, Yifan Jiang, et~al.
\newblock Columbus: Evaluating cognitive lateral understanding through multiple-choice rebuses.
\newblock {\em arXiv:2409.04053}, 2024.

\bibitem[\protect\citeauthoryear{Liang \bgroup \em et al.\egroup }{2024}]{liang-etal-2024-encouraging}
Tian Liang, Zhiwei He, et~al.
\newblock Encouraging divergent thinking in large language models through multi-agent debate.
\newblock In {\em EMNLP}, pages 17889--17904, 2024.

\bibitem[\protect\citeauthoryear{Long \bgroup \em et al.\egroup }{2025}]{puzzleplex}
Yitao Long, Yuru Jiang, et~al.
\newblock Puzzleplex: Benchmarking foundation models on reasoning and planning with puzzles.
\newblock {\em arXiv:2510.06475}, 2025.

\bibitem[\protect\citeauthoryear{Lyu \bgroup \em et al.\egroup }{2025}]{jigsaw-puzzles}
Zesen Lyu, Dandan Zhang, et~al.
\newblock Jigsaw-puzzles: From seeing to understanding to reasoning in vision-language models.
\newblock {\em arXiv:2505.20728}, 2025.

\bibitem[\protect\citeauthoryear{Mahajan \bgroup \em et al.\egroup }{2025}]{mahajan2025multitokenpredictionpretrainingllms}
Divyat Mahajan, Sachin Goyal, et~al.
\newblock Beyond multi-token prediction: Pretraining llms with future summaries.
\newblock {\em arXiv:2510.14751}, 2025.

\bibitem[\protect\citeauthoryear{Ma\l{}ki\'{n}ski and Ma\'{n}dziuk}{2025}]{ai-raven-mesh}
Miko\l{}aj Ma\l{}ki\'{n}ski and Jacek Ma\'{n}dziuk.
\newblock A-i-raven and i-raven-mesh: two new benchmarks for abstract visual reasoning.
\newblock In {\em IJCAI}, 2025.

\bibitem[\protect\citeauthoryear{Markaki and Panagiotakis}{2023}]{Markaki2023JigsawSurvey}
Smaragda Markaki and Costas Panagiotakis.
\newblock Jigsaw puzzle solving techniques and applications: A survey.
\newblock {\em The Vis. Comp.}, 39:4405--4421, 2023.

\bibitem[\protect\citeauthoryear{Małkiński \bgroup \em et al.\egroup }{2025}]{małkiński2025reasoninglimitationsmultimodallarge}
Mikołaj Małkiński, Szymon Pawlonka, and Jacek Mańdziuk.
\newblock Reasoning limitations of multimodal large language models. a case study of bongard problems.
\newblock {\em arXiv:2411.01173}, 2025.

\bibitem[\protect\citeauthoryear{Opielka \bgroup \em et al.\egroup }{2024}]{Opielka2024DoLL}
Gustaw Opielka, Hannes Rosenbusch, Veerle Vijverberg, and Claire~E. Stevenson.
\newblock Do large language models solve arc visual analogies like people do?
\newblock {\em arXiv}, abs/2403.09734, 2024.

\bibitem[\protect\citeauthoryear{Paglieri \bgroup \em et al.\egroup }{2025}]{balrog}
Davide Paglieri, Bart{\l}omiej Cupia{\l}, et~al.
\newblock {BALROG}: Benchmarking agentic {LLM} and {VLM} reasoning on games.
\newblock In {\em ICLR}, 2025.

\bibitem[\protect\citeauthoryear{Pawlonka \bgroup \em et al.\egroup }{2025}]{bongardpwr}
Szymon Pawlonka, Mikołaj Małkiński, and Jacek Mańdziuk.
\newblock Bongard-rwr+: Real-world representations of fine-grained concepts in bongard problems.
\newblock {\em arXiv:2508.12026}, 2025.

\bibitem[\protect\citeauthoryear{Pham \bgroup \em et al.\egroup }{2025}]{iqbench}
Tan-Hanh Pham, Phu-Vinh Nguyen, et~al.
\newblock Iqbench: How "smart" are vision-language models? a study with human iq tests.
\newblock {\em ArXiv:2505.12000}, 2025.

\bibitem[\protect\citeauthoryear{Raven and Raven}{2003}]{rpm}
John Raven and Jean Raven.
\newblock Raven progressive matrices.
\newblock In {\em Handbook of Nonverbal Assessment}, pages 223--237. Springer, Boston, MA, 2003.

\bibitem[\protect\citeauthoryear{Ren \bgroup \em et al.\egroup }{2025}]{vgrp-bench}
Yufan Ren, Konstantinos Tertikas, et~al.
\newblock Vgrp-bench: Visual grid reasoning puzzle benchmark for large vision-language models.
\newblock {\em arXiv:2503.23064}, 2025.

\bibitem[\protect\citeauthoryear{Seely \bgroup \em et al.\egroup }{2025}]{sudokubench}
Jeffrey Seely, Yuki Imajuku, et~al.
\newblock Sudoku-bench: Evaluating creative reasoning with sudoku variants.
\newblock {\em arXiv:2505.16135}, 2025.

\bibitem[\protect\citeauthoryear{Shojaee \bgroup \em et al.\egroup }{2025}]{illusion}
Parshin Shojaee, Seyed~Iman Mirzadeh, Keivan Alizadeh, et~al.
\newblock The illusion of thinking: Understanding the strengths and limitations of reasoning models via the lens of problem complexity.
\newblock In {\em NeurIPS}, 2025.

\bibitem[\protect\citeauthoryear{Song \bgroup \em et al.\egroup }{2025}]{visualpuzzles}
Yueqi Song, Tianyue Ou, et~al.
\newblock Visualpuzzles: Decoupling multimodal reasoning evaluation from domain knowledge.
\newblock {\em arXiv:2504.10342}, 2025.

\bibitem[\protect\citeauthoryear{Tang \bgroup \em et al.\egroup }{2025}]{lego-puzzles}
Kexian Tang, Junyao Gao, et~al.
\newblock Lego-puzzles: How good are mllms at multi-step spatial reasoning?
\newblock {\em arXiv:2503.19990}, 2025.

\bibitem[\protect\citeauthoryear{Todd \bgroup \em et al.\egroup }{2024}]{Todd2024MissedCL}
Graham Todd, Timothy Merino, et~al.
\newblock Missed connections: Lateral thinking puzzles for large language models.
\newblock {\em 2024 IEEE Conference on Games (CoG)}, pages 1--8, 2024.

\bibitem[\protect\citeauthoryear{Unsal and Akkus}{2025}]{easyarc}
Mert Unsal and Aylin Akkus.
\newblock Easyarc: Evaluating vision language models on true visual reasoning.
\newblock {\em arXiv:2506.11595}, 2025.

\bibitem[\protect\citeauthoryear{Wang \bgroup \em et al.\egroup }{2025a}]{enigmaeval}
Clinton~J. Wang, Dean Lee, et~al.
\newblock Enigmaeval: A benchmark of long multimodal reasoning challenges.
\newblock {\em arXiv:2502.08859}, 2025.

\bibitem[\protect\citeauthoryear{Wang \bgroup \em et al.\egroup }{2025b}]{vlm-game-players}
Xinyu Wang, Bohan Zhuang, and Qi~Wu.
\newblock Are large vision language models good game players?
\newblock {\em arXiv:2503.02358}, 2025.

\bibitem[\protect\citeauthoryear{Weng \bgroup \em et al.\egroup }{2025}]{geosketch}
Shichao Weng, Zhiqiang Wang, et~al.
\newblock Geosketch: A neural-symbolic approach to geometric multimodal reasoning with auxiliary line construction and affine transformation.
\newblock {\em arXiv:2509.22460}, 2025.

\bibitem[\protect\citeauthoryear{Wu \bgroup \em et al.\egroup }{2024}]{bongardopenworld}
Rujie Wu, Xiaojian Ma, et~al.
\newblock Bongard-openworld: Few-shot reasoning for free-form visual concepts in the real world.
\newblock In {\em ICLR}, 2024.

\bibitem[\protect\citeauthoryear{W{\"u}st \bgroup \em et al.\egroup }{2025}]{bongard-in-wonderland}
Antonia W{\"u}st, Tim Tobiasch, et~al.
\newblock Bongard in wonderland: Visual puzzles that still make ai go mad?
\newblock In {\em ICML}, 2025.

\bibitem[\protect\citeauthoryear{Xiao \bgroup \em et al.\egroup }{2024}]{logicvista}
Yijia Xiao, Edward Sun, et~al.
\newblock Logicvista: Multimodal llm logical reasoning benchmark in visual contexts.
\newblock {\em arXiv:2407.04973}, 2024.

\bibitem[\protect\citeauthoryear{Xu \bgroup \em et al.\egroup }{2025}]{visulogic}
Weiye Xu, Jiahao Wang, et~al.
\newblock Visulogic: A benchmark for evaluating visual reasoning in multi-modal large language models.
\newblock {\em arXiv:2504.15279}, 2025.

\bibitem[\protect\citeauthoryear{Yan \bgroup \em et al.\egroup }{2025}]{visuriddles}
Hao Yan, Handong Zheng, et~al.
\newblock Visuriddles: Fine-grained perception is a primary bottleneck for multimodal large language models in abstract visual reasoning.
\newblock {\em ArXiv:2506.02537}, 2025.

\bibitem[\protect\citeauthoryear{Yilmaz \bgroup \em et al.\egroup }{2025}]{voila}
Nilay Yilmaz, Maitreya Patel, et~al.
\newblock Voila: Evaluation of {MLLM}s for perceptual understanding and analogical reasoning.
\newblock In {\em ICLR}, 2025.

\bibitem[\protect\citeauthoryear{Yuan \bgroup \em et al.\egroup }{2025}]{mme-reasoning}
Jiakang Yuan, Tianshuo Peng, et~al.
\newblock Mme-reasoning: A comprehensive benchmark for logical reasoning in mllms.
\newblock {\em arXiv:2505.21327}, 2025.

\bibitem[\protect\citeauthoryear{Zhang \bgroup \em et al.\egroup }{2024a}]{ingvp}
Haoran Zhang, Hangyu Guo, et~al.
\newblock Ing-vp: Mllms cannot play easy vision-based games yet.
\newblock {\em arXiv:2410.06555}, 2024.

\bibitem[\protect\citeauthoryear{Zhang \bgroup \em et al.\egroup }{2024b}]{zhang2024how}
Yizhe Zhang, Richard~He Bai, et~al.
\newblock How far are we from intelligent visual deductive reasoning?
\newblock In {\em COLM}, 2024.

\bibitem[\protect\citeauthoryear{Zhang \bgroup \em et al.\egroup }{2025a}]{zhang2025thinkvisuallyreasontextually}
Beichen Zhang, Yuhang Zang, et~al.
\newblock Think visually, reason textually: Vision-language synergy in arc.
\newblock {\em arXiv:2511.15703}, 2025.

\bibitem[\protect\citeauthoryear{Zhang \bgroup \em et al.\egroup }{2025b}]{zhang2025mmcotabenchmarkprobingvisual}
Jusheng Zhang, Kaitong Cai, Xiaoyang Guo, et~al.
\newblock Mm-cot:a benchmark for probing visual chain-of-thought reasoning in multimodal models.
\newblock {\em arXiv:2512.08228}, 2025.

\bibitem[\protect\citeauthoryear{Zhang \bgroup \em et al.\egroup }{2025c}]{puzzlebench}
Zeyu Zhang, Zijian Chen, et~al.
\newblock Puzzlebench: A fully dynamic evaluation framework for large multimodal models on puzzle solving.
\newblock {\em arXiv:2504.10885}, 2025.

\bibitem[\protect\citeauthoryear{Zhou \bgroup \em et al.\egroup }{2025a}]{zhou2025perceptioncognitionsurveyvisionlanguage}
Chenyue Zhou, Mingxuan Wang, et~al.
\newblock From perception to cognition: A survey of vision-language interactive reasoning in multimodal large language models.
\newblock {\em arXiv:2509.25373}, 2025.

\bibitem[\protect\citeauthoryear{Zhou \bgroup \em et al.\egroup }{2025b}]{manbench}
Han Zhou, Qitong Xu, et~al.
\newblock {MANB}ench: Is your multimodal model smarter than human?
\newblock In {\em ACL Findings}, pages 3423--3449, 2025.

\end{thebibliography}

\end{document}